\documentclass[runningheads]{llncs}
\usepackage{booktabs}
\usepackage{multirow}
\usepackage[colorlinks]{hyperref}
\usepackage{graphicx}
\usepackage[misc]{ifsym}
\begin{document}
\title{A Multi-task Framework for Skin Lesion Detection and Segmentation   
}
%
%
\author{Sulaiman Vesal$^*$\inst{1}(\Letter)\and Shreyas Malakarjun Patil$^*$\inst{1,2}(\Letter)\and
 Nishant Ravikumar\inst{1} \and
 Andreas K. Maier\inst{1}}

\authorrunning{S. Vesal et al.}

\institute{Pattern Recognition Lab, Friedrich-Alexander-Universit\"at Erlangen-N\"urnberg\, Germany\\
\and Department of Electrical Engineering, Indian Institute of Technology Jodhpur, Rajasthan, India\\
\email{sulaiman.vesal@fau.de, patil.3@iitj.ac.in}\\$^*$ These authors contributed equally to this article}

\maketitle              
\begin{abstract}
Early detection and segmentation of skin lesions is crucial for timely diagnosis and treatment, necessary to improve the survival rate of patients. However, manual delineation is time consuming and subject to intra- and inter-observer variations among dermatologists. This underlines the need for an accurate and automatic approach to skin lesion segmentation. To tackle this issue, we propose a multi-task convolutional neural network (CNN) based, joint detection and segmentation framework, designed to initially localize the lesion and subsequently, segment it. A `Faster region-based convolutional neural network' (Faster-RCNN) which comprises a region proposal network (RPN), is used to generate bounding boxes/region proposals, for lesion localization in each image. The proposed regions are subsequently refined using a softmax classifier and a bounding-box regressor. The refined bounding boxes are finally cropped and segmented using `SkinNet', a modified version of U-Net. We trained and evaluated the performance of our network, using the ISBI 2017 challenge and the PH2 datasets, and compared it with the state-of-the-art, using the official test data released as part of the challenge for the former. Our approach outperformed others in terms of Dice coefficients ($>0.93$), Jaccard index ($>0.88$), accuracy ($>0.96$) and sensitivity ($>0.95$), across five-fold cross validation experiments. 

 
\end{abstract}

\section{Introduction}

Recent trends indicate a growing number of skin cancer diagnoses worldwide, each year. In 2016, approximately 80,000 new cases of skin cancer were expected to be diagnosed, with 10,000 melanoma related deaths  (the most aggressive form of skin cancer), in the USA alone \cite{siegel16}. Clinical screening and diagnosis typically involve examination by an expert dermatologist, followed by histopathological analysis of biopsies. These steps however, invariably suffer from high inter-rater and inter-center variability, and studies have shown that patient survival rates improve to over 95$\%$, following early detection and diagnosis of melanomas. To reduce variability in the screening process, computer-aided-diagnosis (CAD) systems, which enable automatic detection, lesion segmentation and classification of dermoscopic images, in a manner robust to variability in image quality and lesion appearance, are essential. 

Segmentation is an essential initial step, for CAD of skin lesions \cite{mirzaalian2018detecting} and melanoma in particular. This is because melanoma is typically diagnosed based on the `ABCD' criterion, which takes into account the shape-characteristics of lesions (such as diameter, asymmetry, border irregularity, etc.), together with appearance, or the `seven-point checklist' \cite{jafari16}. Consequently, the quality of the initial segmentation is crucial to the subsequent evaluation of diagnostic metrics such as border irregularity and lesion diameter. Several deep learning-based approaches have been proposed, for skin lesion segmentation in recent years, for example - a multi-task CNN was formulated in \cite{yang17}, which simultaneously tackled lesion segmentation and two independent binary classification tasks; the winners of the ISBI 2016 skin lesion segmentation challenge \cite{Yu}, employed a fully convolutional residual network (FCRN), with more than 50 layers for segmentation and integrated it within a 2-stage framework for melanoma classification; and in \cite{kawahara18}, a multi-modal, multi-task CNN was designed, for the classification of the seven-point melanoma checklist criteria, and skin lesion diagnosis.

We proposed a CNN-based segmentation framework called `SkinNet' \cite{vesal18} recently, to segment skin lesions in dermoscopic images automatically. The proposed CNN architecture was a modified version of the U-Net \cite{Unett}. SkinNet employs dilated convolutions in the lowest layer of the encoder-branch, to provide a more global context for the features extracted in the image. Additionally, the model replaced the conventional convolution layers in both the encoder and decoder branches of U-Net, with dense convolution blocks, to better incorporate multi-scale image information. 

In this paper, we propose a novel two-stage approach for skin lesion detection and segmentation where we first localize the lesion, and subsequently segment it. The recently developed `faster region-based convolutional neural network' (Faster-RCNN) \cite{FRCNN}, a form of multi-task learning, is utilized for lesion localization. For each image, a number of bounding-boxes are initially generated by a region proposal network (RPN). Subsequently, each proposed region is jointly classified (as containing the object of interest or not) and refined using a softmax classifier, and a bounding-box regressor. Following refinement, the detected regions are cropped and segmented using SkinNet. 

\section{Methods}
\label{sec:methods}
A fully automatic CAD system for analyzing dermoscopic images, must first be able to accurately localize, and segment the lesion, prior to classifying it into its sub-types. The framework devised in this study for skin lesion segmentation comprises, an initial localization step, using a network designed for object detection, followed by segmentation using a modified U-Net. The overall network was trained using the ISBI 2017 challenge (training) dataset \cite{ISBI}.

\begin{figure}[!ht]
\centering
	\includegraphics[width=10cm]{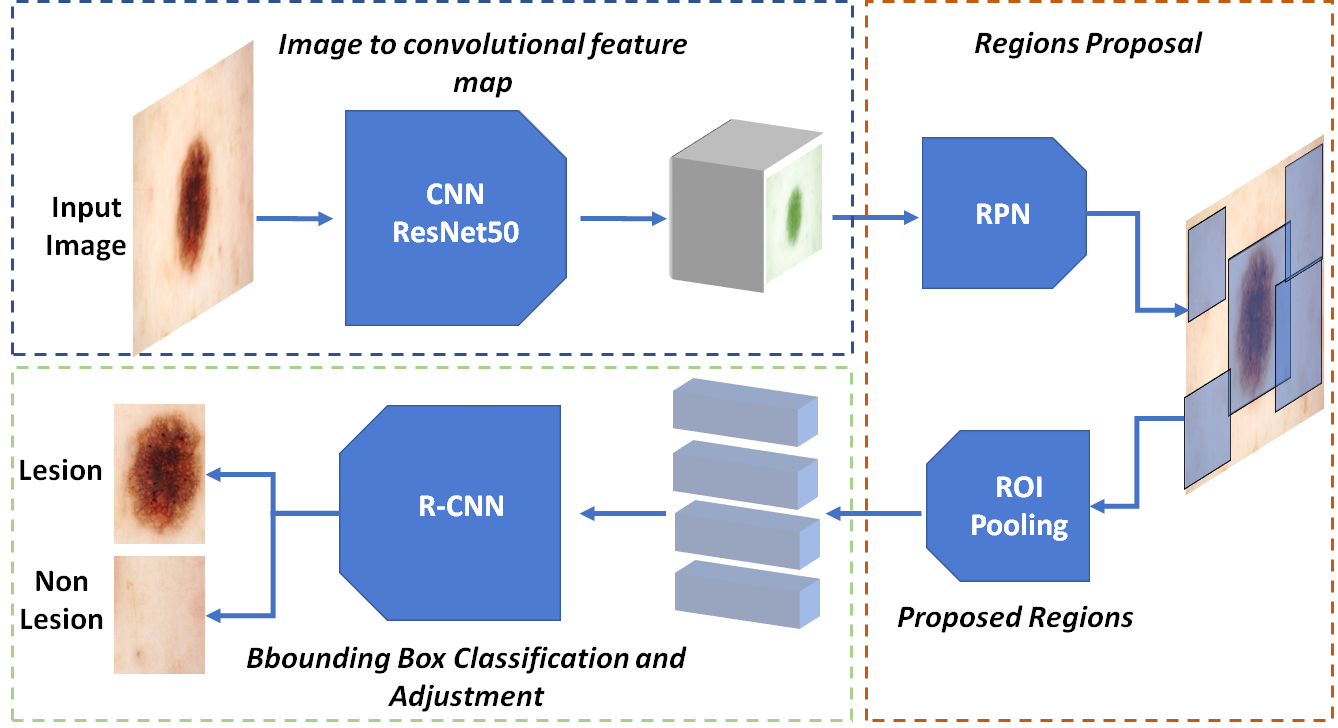}
\caption{Faster-RCNN architecture: Top left box represents the base network, box on the right represents the region proposal network (RPN) and the bottom left box represents the RCNN.} \label{fig1}
\end{figure}

A network similar to the original Faster-RCNN was constructed for the initial task of lesion localization. The network's main components are summarized in Fig. \ref{fig1}. These include: a) shared convolution layers (henceforth referred to as the base network) to extract both low- and high-level features from the input image; b) a region proposal network (RPN)\cite{FRCNN}, which predicts anchor boxes and the probability that the predicted box contains a lesion; and (c) a region-based convolution network (RCNN) which refines the regions of interest (ROIs) generated in the preceding RPN step, by predicting the class (lesion present vs absent), and bounding box coordinates. Following localization, and selection of the refined regions, lesions were segmented within the estimated bounding boxes, using SkinNet. Henceforth, we refer to the combined localization and segmentation framework proposed in this study as, Faster-RCNN+SkinNet.

\textbf{The base network:} In order to extract discriminative features within the shared layers, we employed the pre-trained (on ImageNet) ResNet50 residual network \cite{he16}. The network was split into two parts, the first comprising the initial 87 layers was used as the base network, and the remaining layers were used for classification and regression in the final RCNN (refer to Fig. \ref{fig1}). The 87 layers were chosen based on experiments wherein, the number of layers of the base network were varied. Each trial was evaluated in terms of the Intersection-over-Union (IoU) of the bounding boxes predicted by the Faster-RCNN for each image, with respect to their ground truths, resulting in the chosen configuration. 

\textbf{Region proposal network:} Following feature extraction, nine anchor boxes of various scales and aspect ratios were generated, centered on distinct, non-overlapping $3\times3$ patches of the feature map obtained from the base network, for each image. These anchors were generated at scales of $[128, 256, 512]$, and aspect ratios of $[1:1, 1:2, 2:1]$. The RPN was designed to predict the coordinates of these anchors for all patches, and their probability of containing a lesion. The similarity between the anchor boxes and the ground truth bounding boxes (generated using the training masks provided) was measured using IoU, and used to create references used by the RPN (as synthetic ground truths) to predict the probability of the anchors containing a lesion. These anchor boxes were labeled as positive, negative or neutral, based on IoU thresholds of $0.7$ and $0.4$, respectively. We ensured that the ground truth bounding boxes each had at least one corresponding positive anchor box, and if not, the neutral anchor box with the highest IoU was labeled positive. The RPN was implemented as a set of convolution layers, where each anchor box was first convolved with a $3\times3$ kernel, and subsequently, with five $1\times1$ kernels, resulting in five feature maps. Each of these feature maps in turn represent the coordinates of each anchor box, and its probability of containing a lesion. This process was repeated nine times, for each of the nine types of anchor boxes we considered, resulting in $9\times5$ feature maps that were predicted per image.  

\textbf{Classification and bounding box regression:} Classification of each region proposed by the RPN required feature maps of fixed sizes, as input to the RCNN. These were generated using region of interest (ROI) pooling. During ROI pooling, each feature map from the RPN was cropped and resized to $14\times14\times1024$ via bilinear interpolation. Next, max pooling with a $2\times2$ kernel was used, resulting in a final $7\times7\times1024$ feature map for each proposal. Finally, we used the remaining layers of the ResNet50 architecture (excluded in the base network), implemented as time-distributed layers, for the RCNN. Time-distributed convolution layers were used to avoid iterative classification and regression training and to accommodate the varied number of regions proposed per image, by the RPN. 
The RCNN subsequently classifies each proposal as lesion/non-lesion, and adjusts the bounding box coordinates to fit the lesion completely. Non-Maximum suppression with a threshold of 0.5 was used as a final step, to remove redundant bounding boxes.

\textbf{Skin lesion segmentation:} The final set of ROIs estimated for each image, using the Faster-RCNN based localization network, are subsequently, used as inputs for segmentation, by SkinNet \cite{vesal18} which we proposed in our recent studies. This segmentation network was designed to incorporate both local and global information, beneficial for any segmentation task. In segmentation networks such as the U-Net, the lowest level of the network connecting the encoder and decoder branches, has a small receptive field, which prevents the network from extracting features that capture non-local image information. We addressed this issue by using dilated convolution layers in the lowest part of the network. The encoded features are convolved with successively increasing dilation rates, which in turn, successively increases the size of the receptive field. The encoder and decoder branches of SkinNet each comprise, three down- and up-sampling dense convolution blocks. These blocks incorporate multi-scale information through the use of dense convolution layers, where, the input to every layer is a concatenation of output feature maps, from all preceding convolution layers.

\textbf{Losses:} The losses used for RPN and RCNN classification are cross-entropy, and categorical cross-entropy, respectively. Mean squared error (MSE) was used as the regression loss in both the RPN and the RCNN. The ground truth for the bounding box regression was generated manually using the binary masks provided in the training dataset, for the ISBI 2017 challenge \cite{ISBI}. Many traditional segmentation networks employ cross-entropy \cite{Unett} as a loss function. However, due to the small size of the lesion in dermoscopy images, cross-entropy is biased towards the background of the image. Consequently, for SkinNet, we used a dice coefficient loss function $\zeta(y, \hat{y}) = \zeta(y, \hat{y})  = 1- \sum_{k}\frac{\sum_{n}y_{nk} \hat{y}_{nk}}{\sum_{n}y_{nk} + \sum_{n}\hat{y}_{nk}}$. The dice loss was chosen as experimental evidence suggested that it is less affected by class imbalances.
Here, $\hat{y}_{nk}$ denotes the output of the model, where $n$ represents the pixels and $k$ the classes (i.e. background vs. lesion). The ground truth masks are one-hot encoded and denoted by $y_{nk}$. We take one minus the dice coefficient in order to constrain the loss to zero.
\begin{figure}[tb]
\centering
\includegraphics[width=10cm]{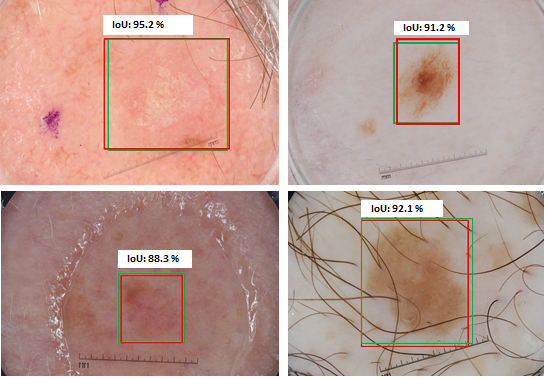}
\caption{Some examples of detected lesions and their respective IoU scores. The green and red bounding boxes represent the ground truth and predicted boxes, respectively.} \label{fig2}
\end{figure}

\textbf{Training procedure:} A four-step training process for each batch was used in our approach. In the first step, we trained the RPN for a batch, generating numerous region proposals. Subsequently, the classification and bounding box regression branches of the RCNN were trained for the same batch. During both these steps, the weights of the base network were also fine tuned to enable the network to learn task specific features. Next, the weights of the base network were frozen and the RPN was fine tuned, to predict the anchor boxes. Finally, the classification and regression branches of the RCNN were also fine tuned, once again keeping the weights of the base network fixed. The proposed detection method was trained for 100 epochs, using the Adam optimizer with a learning rate of 0.001. The model achieved an accuracy of 95.0\% on the validation set ($20\%$ of the training set) and 94.0\% on the test set ($10\%$ of the training set) respectively, for an overlap threshold of $0.9$. Example outputs of lesion detection on test data are depicted in Fig \ref{fig2}, which clearly highlight the high detection accuracy of the proposed approach.





\section{Results and Discussion}
\label{sec:res}
\textbf{Datasets:} In order to evaluate the performance of our approach, we trained and tested it on two well-known public datasets, namely, the ISBI 2017 challenge dataset \cite{ISBI} and the PH2 \cite{mendoncca13} dataset. The former includes 2000 dermoscopic images and their corresponding lesion masks. These images are of various dimensions ranging from $1022\times767$ to $6688\times4439$. In addition to the training set, the organizers also provided a validation set comprising 150 images, and an additional test set with 600 images for final evaluation. The PH2 dataset contains 200 images, each $786\times560$ in size, and acquired at a magnification of $20\times$. We used these images purely as unseen data, to test the ability of our framework to generalize to images obtained from a different database. All images were resized to 512$\times$512$\times$3. The number of images from both datasets used for training, validation and testing, are summarized in Table \ref{tab:tab1}.

\begin{table}[tb]
  \centering
  \caption{Distribution of the ISBI 2017 challenge and PH2 datasets.}
    \begin{tabular}{p{6.43em}p{7.145em}p{8.5em}cc}
    \toprule
    \textbf{Dataset} & \textbf{Training Data} & \textbf{Validation Data} & \multicolumn{1}{p{6.07em}}{\textbf{Test Data}} & \multicolumn{1}{p{4.715em}}{\centering\textbf{Total}} \\
\cmidrule{2-4}    ISBI2017  & \multicolumn{1}{c}{2000} & \multicolumn{1}{c}{150} & 600       & 2750 \\
    PH2       & \centering-         & \centering-         & 200       & 200 \\
    \bottomrule
    \end{tabular}%
  \label{tab:tab1}%
\end{table}%
\begin{table}[tb]
  \centering
  \caption{The segmentation accuracy results for different methods on ISBI 2017 challenge test data.}
    \begin{tabular}{lp{8.965em}ccccc}
    \toprule
    \multicolumn{1}{p{4.235em}}{\textbf{Datasets}} & \textbf{Methods} & \multicolumn{1}{p{4.235em}}{\centering\textbf{AC}} & \multicolumn{1}{p{4.235em}}{\centering\textbf{DC}} & \multicolumn{1}{p{4.235em}}{\centering\textbf{JI}} & \multicolumn{1}{p{4.235em}}{\centering\textbf{SE}} & \multicolumn{1}{p{4.235em}}{\centering\textbf{SP}} \\
    \midrule
    \multicolumn{1}{l}{\multirow{6}[3]{*}{\textbf{ISBI2017}}} & Yuan et. al. \cite{Yuan} & 0.934     & 0.849     & 0.765     & 0.825     & 0.975 \\
              & SLSDeep \cite{SLICNet}   & 0.936     & 0.878     & 0.782     & 0.816     & 0.983 \\
              & NCARG \cite{sym10040119}     & 0.953     & 0.904     & 0.832     & 0.975     & 0.888 \\
              & FrCN \cite{ALMASNI2018221}      & 0.956     & 0.896     & 0.813     & 0.890     & 0.974 \\
            & \textbf{SkinNet} & 0.932     & 0.851     & 0.767     & 0.930     & 0.905 \\
              & \textbf{Faster-RCNN+SkinNet} & \textbf{0.968} & \textbf{0.934} & \textbf{0.880} & 0.971 & 0.913 \\
 \cmidrule{2-7}
              & FrCN \cite{ALMASNI2018221}      & 0.952 & 0.914 & 0.841 & 0.945 & 0.955 \\
    \multicolumn{1}{l}{\multirow{2}[2]{*}{\textbf{PH2}}} & \textbf{Faster-RCNN+SkinNet} & \textbf{0.964} & \textbf{0.946} & \textbf{0.899} & \textbf{0.952} & 0.925 \\
    \bottomrule
    \end{tabular}%
  \label{tab:addlabel}%
\end{table}%
\begin{figure}[h]
\centering
\includegraphics[width=10cm]{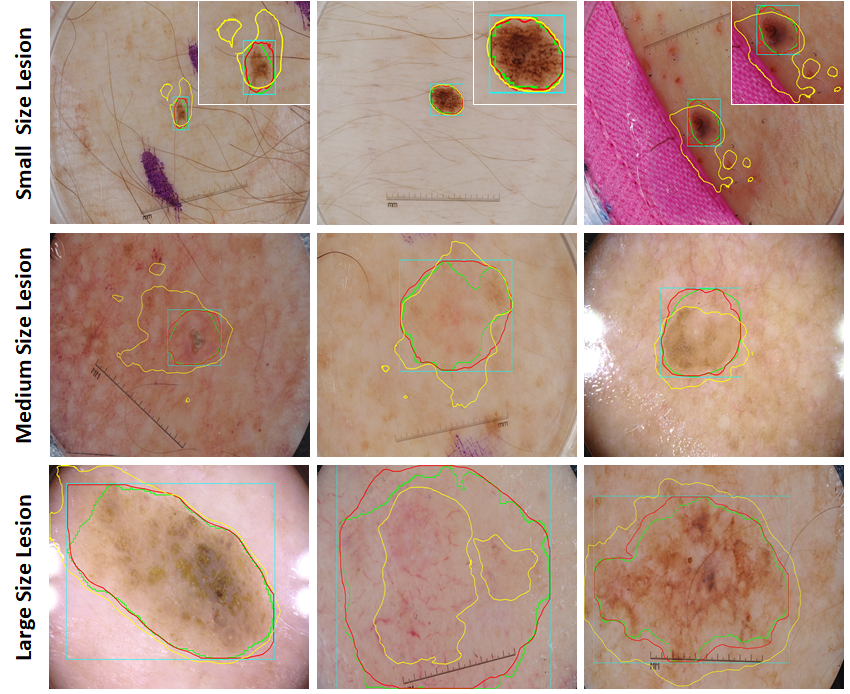}
\caption{Segmentation outputs using SkinNet and Faster-RCNN+SkinNet for different lesion sizes. The blue rectangle represents the detected bounding box. The green contour represents the ground truth segmentation, while the red and yellow represent the outputs of Faster-RCNN+SkinNet and SkinNet, respectively. } \label{fig3}

\end{figure}
\textbf{Evaluation Metrics:} We used the metrics employed in the ISBI 2017 challenge, to evaluate segmentation performance, namely, Specificity(SP), Sensitivity(SE), Jaccard index(JI), Dice coefficient(DC) and Accuracy(AC), across five-fold cross validation experiments. Table \ref{tab:tab1} summarizes segmentation accuracy, evaluated using each of these metrics, for SkinNet and Faster-RCNN+SkinNet, on the ISBI 2017 test set and the PH2 data set. It also compares the achieved results with the state-of-the-art, which were trained and tested on the same data. For the ISBI 2017 test data, Faster-RCNN+SkinNet outperformed SkinNet and all other methods in terms of AC, DC, JI and SE. In particular, it achieved an average DC and JI score of 93.4\% and 88\%, respectively, which is significantly higher than all other methods. Visual assessment of the segmentation accuracy of Faster-RCNN+SkinNet relative to SkinNet, depicted in Fig.\ref{fig3}, confirms the superiority of the former relative to the latter. Furthermore, for the PH2 dataset, our method once again outperformed a state-of-the-art approach \cite{ALMASNI2018221}, in terms of AC, DC, JI and SE, highlighting its ability to generalize to images acquired from other databases. These results and comparisons, clearly outline the improvement in segmentation accuracy achieved by the proposed approach, relative to the state-of-the-art, and by extension, the benefit of formulating a multi-task learning approach, for skin lesion segmentation.

\section{Conclusion}
The multi-task framework proposed in this study for joint lesion localization and segmentation, significantly outperformed the state-of-the-art, on two public test data sets. The results outline the significant benefits of object localization and multi-task learning, as auxiliaries to segmentation tasks. The proposed framework thus shows promise for the automatic analysis of skin lesions in dermoscopic images, for improved diagnosis and clinical decision support.
\label{sec:conclude}

\section*{Acknowledgements}
This study was partially supported by the project - BIG-THERA: Integrative `Big Data Modeling' for the development of novel therapeutic approaches for breast cancer.
\bibliographystyle{splncs03}

\bibliography{0000}

\begin{thebibliography}{10}

\bibitem{siegel16}
Siegel, R.L., Miller, K.D., Jemal, A.:
\newblock Cancer statistics, 2016.
\newblock CA: a cancer journal for clinicians \textbf{66}(1) (2016)  7--30

\bibitem{mirzaalian2018detecting}
Mirzaalian-Dastjerdi, H., T{\"o}pfer, D., Bangemann, M., Maier, A.:
\newblock Detecting and measuring surface area of skin lesions.
\newblock In: Bildverarbeitung f{\"u}r die Medizin 2018.
\newblock Springer (2018)  29--34

\bibitem{jafari16}
Jafari, M.H., Karimi, N., Nasr-Esfahani, E., Samavi, S., Soroushmehr, S.M.R.,
  Ward, K., Najarian, K.:
\newblock Skin lesion segmentation in clinical images using deep learning.
\newblock In: Pattern Recognition (ICPR), 2016 23rd International Conference
  on, IEEE (2016)  337--342

\bibitem{yang17}
Yang, X., Zeng, Z., Yeo, S.Y., Tan, C., Tey, H.L., Su, Y.:
\newblock A novel multi-task deep learning model for skin lesion segmentation
  and classification.
\newblock arXiv preprint arXiv:1703.01025 (2017)

\bibitem{Yu}
Yu, L., Chen, H., Dou, Q., Qin, J., Heng, P.A.:
\newblock Automated melanoma recognition in dermoscopy images via very deep
  residual networks.
\newblock IEEE Transactions on Medical Imaging \textbf{36}(4) (April 2017)
  994--1004

\bibitem{kawahara18}
Kawahara, J., Daneshvar, S., Argenziano, G., Hamarneh, G.:
\newblock 7-point checklist and skin lesion classification using multi-task
  multi-modal neural nets.
\newblock IEEE Journal of Biomedical and Health Informatics (2018)

\bibitem{vesal18}
Vesal, S., Ravikumar, N., Maier, A.:
\newblock Skinnet: A deep learning framework for skin lesion segmentation.
\newblock (2018) preprint, \url{https://arxiv.org/abs/1806.09522}.

\bibitem{Unett}
Ronneberger, O., Fischer, P., Brox, T.:
\newblock U-net: Convolutional networks for biomedical image segmentation.
\newblock In Navab, N., Hornegger, J., Wells, W.M., Frangi, A.F., eds.: Medical
  Image Computing and Computer-Assisted Intervention -- MICCAI 2015, Cham,
  Springer International Publishing (2015)  234--241

\bibitem{FRCNN}
Ren, S., He, K., Girshick, R., Sun, J.:
\newblock Faster r-cnn: Towards real-time object detection with region proposal
  networks.
\newblock In: Advances in neural information processing systems. (2015)  91--99

\bibitem{ISBI}
Codella, N.C.F., Gutman, D., Celebi, M.E., Helba, B., Marchetti, M.A., Dusza,
  S.W., Kalloo, A., Liopyris, K., Mishra, N.K., Kittler, H., Halpern, A.:
\newblock Skin lesion analysis toward melanoma detection: {A} challenge at the
  2017 international symposium on biomedical imaging (isbi), hosted by the
  international skin imaging collaboration {(ISIC)}.
\newblock CoRR \textbf{abs/1710.05006} (2017)

\bibitem{he16}
He, K., Zhang, X., Ren, S., Sun, J.:
\newblock Deep residual learning for image recognition.
\newblock In: Proceedings of the IEEE conference on computer vision and pattern
  recognition. (2016)  770--778

\bibitem{mendoncca13}
Mendon{\c{c}}a, T., Ferreira, P.M., Marques, J.S., Marcal, A.R., Rozeira, J.:
\newblock Ph 2-a dermoscopic image database for research and benchmarking.
\newblock In: Engineering in Medicine and Biology Society (EMBC), 2013 35th
  Annual International Conference of the IEEE, IEEE (2013)  5437--5440

\bibitem{Yuan}
Yuan, Y., Chao, M., Lo, Y.C.:
\newblock Automatic skin lesion segmentation using deep fully convolutional
  networks with jaccard distance.
\newblock IEEE Transactions on Medical Imaging \textbf{36}(9) (Sept 2017)
  1876--1886

\bibitem{SLICNet}
Kamal~Sarker, M.M., Rashwan, H.A., Akram, F., Furruka~Banu, S., Saleh, A.,
  Singh, V.K., Chowdhury, F.U.H., Abdulwahab, S., Romani, S., Radeva, P., Puig,
  D.:
\newblock Slsdeep: Skin lesion segmentation based on dilated residual and
  pyramid pooling networks, eprint arXiv:1805.10241 (2018)

\bibitem{sym10040119}
Guo, Y., Ashour, A.S., Smarandache, F.:
\newblock A novel skin lesion detection approach using neutrosophic clustering
  and adaptive region growing in dermoscopy images.
\newblock Symmetry \textbf{10}(4) (2018)

\bibitem{ALMASNI2018221}
Al-masni, M.A., Al-antari, M.A., Choi, M.T., Han, S.M., Kim, T.S.:
\newblock Skin lesion segmentation in dermoscopy images via deep full
  resolution convolutional networks.
\newblock Computer Methods and Programs in Biomedicine \textbf{162} (2018)  221
  -- 231

\end{thebibliography}
\end{document}